\documentclass[11pt,a4paper]{article}

\pdfoutput=1
\usepackage{paralist}

\newcommand{\Sref}[1]{\S\ref{#1}}
\newcommand{\ignore}[1]{}
 
 \usepackage{hyperref}

\usepackage[]{acl}
\usepackage{amsmath}

\usepackage{times}
\usepackage{latexsym}
\usepackage{multirow}
\usepackage{tabularx}
\usepackage[T1]{fontenc}
\usepackage{mathtools}
\usepackage{float}

\usepackage{pgf}
\usepackage{pgfplots}
\pgfplotsset{compat=1.17}

\usepackage[utf8]{inputenc}

\usepackage{microtype}
\usepackage{graphicx}

\definecolor{Blue}{HTML}{0062cc}
\definecolor{ForestGreen}{HTML}{006400}

\title{Speaker Information Can Guide Models to Better Inductive Biases: \\A Case Study On Predicting Code-Switching}

\author{Alissa Ostapenko$^1$ \And Shuly Wintner$^2$ \And Melinda Fricke$^3$ \And Yulia Tsvetkov$^4$\\ \add $^1$Language Technologies Institute, Carnegie Mellon University\\
\add $^2$Department of Computer Science, University of Haifa\\
\add $^3$Department of Linguistics, University of Pittsburgh\\
\add $^4$Paul G. Allen School of Computer Science \& Engineering, University of Washington\\
\email aostapen@andrew.cmu.edu, shuly@cs.haifa.ac.il, \\  \email melinda.fricke@pitt.edu, yuliats@cs.washington.edu  }
  
\begin{document}
\maketitle
\begin{abstract}
Natural language processing (NLP) models trained on people-generated data can be unreliable because, without any constraints, they can learn from spurious correlations that are not relevant to the task. We hypothesize that enriching models with speaker information in a controlled, educated way can guide them to pick up on relevant inductive biases. For the speaker-driven task of predicting code-switching points in English--Spanish bilingual dialogues, we show that adding sociolinguistically-grounded speaker features as prepended prompts significantly improves accuracy.  We find that by adding influential phrases to the input, speaker-informed models learn useful and explainable linguistic information. To our knowledge, we are the first to incorporate speaker characteristics in a neural model for code-switching, and more generally, take a step towards developing transparent, personalized models that use speaker information in a controlled way.
\end{abstract}
\setlength{\belowcaptionskip}{-10pt}

\section{Introduction}

Imbalanced datasets, flawed annotation schemes, and even model architectures themselves  can all cause neural models to encode and propagate biases by incorrectly correlating social information with labels for a task  \cite{sun-etal-2019-mitigating,field21}. As a result, models may be brittle and offensive in the presence of racial or gender attributes \cite{kchenko2018, nozza2021honest}, unsuitable for processing mixed-language text or dialect variations \cite{sap-etal-2019-risk, kumar21, winata2021}, or ones that can miscommunicate intents in translation setups. 
Contextualizing models in social factors is important for preventing these issues and building more socially intelligent and culturally sensitive NLP technologies \citep{hovy2021importance}. 

We hypothesize that grounding models in speaker information can help them learn more useful inductive biases, thereby improving performance on person-oriented classification tasks.
We test this hypothesis on the task of predicting code-switching (language mixing) in a multilingual dialogue, which is inherently linguistically \emph{and} socially driven \cite{li-cs}. 
Prior approaches for predicting code-switching consider only shallow linguistic context \cite{dogruoz-etal-2021-survey}. As we show in our experiments\footnote{All data and code will be available at \url{https://github.com/ostapen/Switch-and-Explain}.}, using a standard Transformer-based classifier \cite{conneau-etal-2020-unsupervised}  trained with only linguistic context results in sub-optimal and unstable models. Moreover, we believe code-switch prediction is a suitable first task for learning speaker-driven inductive biases; we can test whether models learn useful relationships between social attributes while minimizing the risk of building a model that perpetuates social prejudices. 

We ground the models in relevant social factors, such as age, native language, and language-mixing preference of the interlocutors, via text-based speaker descriptions or \emph{prompts} \citep[cf.][]{zhong2021adapting,wei2021finetuned}.
We find that prepending speaker prompts to dialogue contexts improves performance significantly, and leads to more stable generalizations. Our prompts are different from the embedding-based personas of \citet{li-etal-2016-persona} and the synthesized descriptions from Persona-Chat \citep{zhang2018personalizing}, capturing theoretically grounded social and linguistic properties of speakers, as opposed to hobbies or occupations. 

To analyze the inductive biases that the models learn, 
we use \textit{SelfExplain} \cite{rajagopal2021selfexplain}---an interpretable  text classification model highlighting key phrases in the input text. 
We propose a new method for aggregating the interpretations produced by SelfExplain to explain model predictions and align them with sociolinguistic literature. 

\ignore{

}
We motivate our study of predicting code-switching in \Sref{sec:motivation}, and  describe the task and interpretable neural text classification models in \Sref{sec:methodology}. After outlining important ethical considerations in \Sref{sec:ethics}, we detail our experiments (\Sref{sec:experimental}) and results (\Sref{sec:eval}), and provide an analysis of speaker-aware model generalizations that are grounded in prior psycholinguistic research on code-switching (\Sref{sec:analysis}).

\section{Motivation}
\label{sec:motivation}
Our overarching goal is to develop a general and theoretically-informed methodology to ground neural models in a social context, because a wide array of person-centric classification tasks, such as sentiment prediction or hate speech detection, can fail without proper social contextualization  \cite{sap-etal-2019-risk, kchenko2018, hovy2021importance}. 
We choose a speaker-driven task that is ethically safer to experiment with (see a detailed discussion in \Sref{sec:ethics}): predicting code-switching in human--human dialogues. 

 \emph{Code-switching} is the alternation between languages within and between utterances (See Appendix \ref{app:cs-example} for a detailed example of code-switched dialogue.) It is a language- and speaker-driven phenomenon, reflecting speaker identities and relationships between them, in addition to their linguistic backgrounds, preferences and topical constraints  \cite{cs-martinez}. 
Prior sociolinguistic work established the importance of speaker context for code-switching, and existing multilingual models---trained with only monolingual linguistic context---are not speaker-grounded nor well-suited for dealing with mixed-language data, leaving gaps which we begin to address.

\begin{figure}[t!]
    \centering
    \includegraphics[clip, trim={38cm 0cm 0cm 0cm}, scale=0.25]{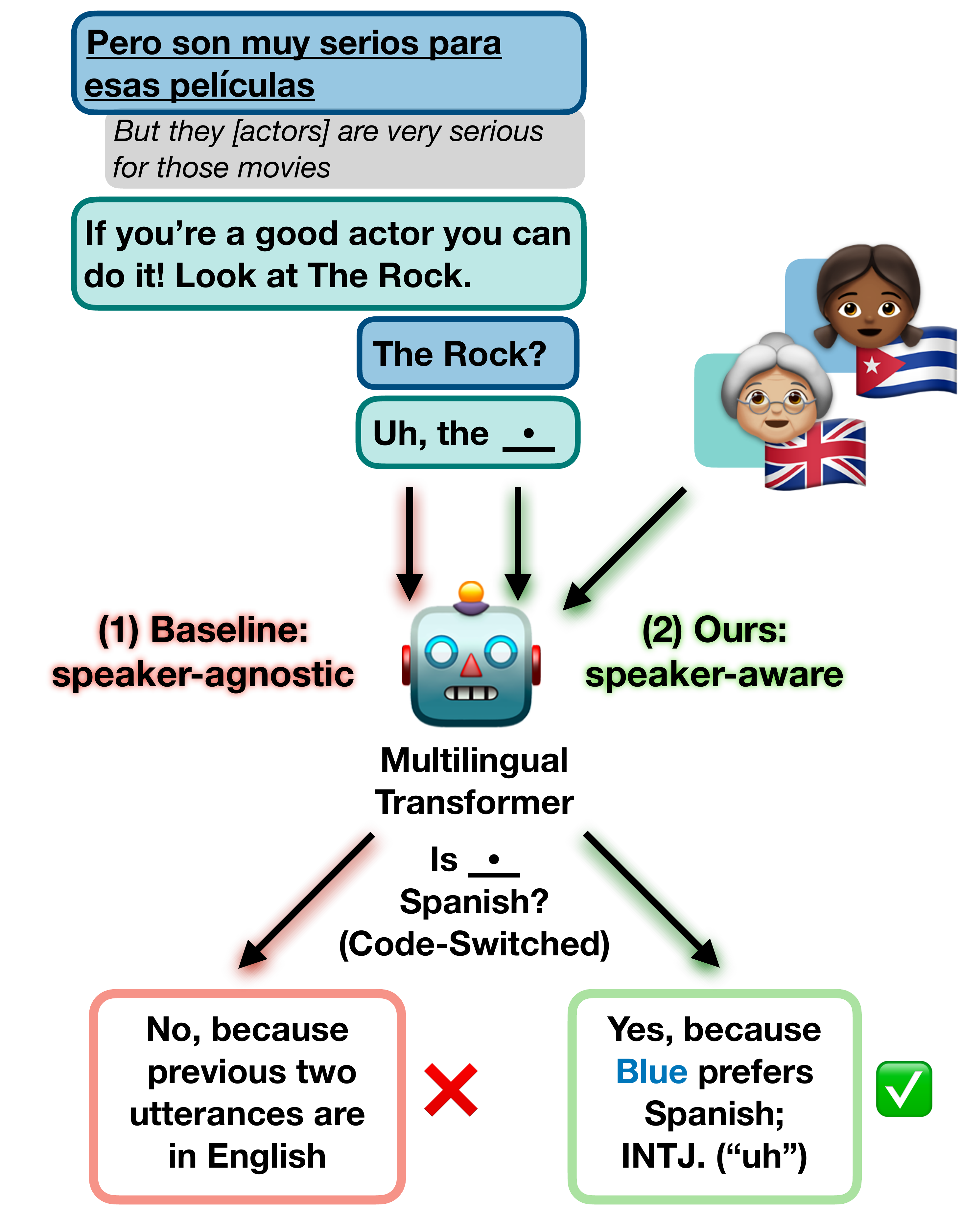}
    \caption{We use a Transformer-based model to predict language switches in dialogues and identify phrase-level features guiding predictions. Here, both speakers are bilingual, but Blue's native language is Spanish and Green's native language is English. They have unique social factors (such as age). The dialogue structure reflects speaker identities and relationships: Green will switch to Spanish with  \textit{el actor}, accommodating Blue's language preference. Using only dialogue context, the baseline (1) fails to pick up on this, while our speaker-aware model (2) successfully predicts a code-switch and identifies useful linguistic cues.}
    \label{fig:cs_example}
\end{figure}

Figure \ref{fig:cs_example} provides a key motivating example of how global speaker features of two bilingual conversational participants influence their local speech production. \textit{Blue}, whose native language is Spanish, begins speaking in Spanish, while \textit{Green} responds in English. Following \textit{Green}'s clarification question about the actor \textit{The Rock}, \textit{Green} begins in English, but will accommodate \textit{Blue} \cite{ahn2019codeswitch, cs-martinez} to reply with \textit{el actor} (Spanish), motivating the need for social context when processing mixed-language data.

\section{Methodology}
\begin{table*}[bth]
\begin{tabularx}\textwidth{l|X}

\hline
\textbf{Prompt} & \textbf{Speaker Description Example} \\
\hline
\multirow{2}{4em}{\textbf{List}}   &  \textcolor{violet}{\textbf{\textsc{ASH}}} is first speaker, older,  female, from Spanish speaking country, between English and Spanish prefers both,  rarely switches languages. \\ [0.55cm] & \textcolor{Blue}{\textbf{\textsc{JAC}}} is second speaker, older,  male, from Spanish speaking country, between English and Spanish prefers both,  never switches languages. \\
\hline
\multirow{2}{4em}{\textbf{Sentence}}   & \textcolor{violet}{\textbf{\textsc{ASH}}} is a  middle-aged  woman  from a Spanish speaking country. Between English and Spanish she prefers both, and she rarely switches languages.  \textcolor{violet}{\textbf{\textsc{ASH}}} speaks first.  \\ [0.55cm] &  \textcolor{Blue}{\textbf{\textsc{JAC}}} is a middle-aged man from a Spanish speaking country. Between English and Spanish he prefers both, and he never switches languages. \textcolor{Blue}{\textbf{\textsc{JAC}}} speaks second.
 \\
 
\hline
\multirow{2}{4em}{\textbf{Partner}} & \textcolor{ForestGreen}{\textbf{\textsc{ASH}, \textsc{JAC}}} are all middle-aged  from a Spanish speaking country. Between English and Spanish they prefer both.    \\[0.55cm] & \textcolor{violet}{\textbf{\textsc{ASH}}} is a woman and rarely switches languages. \textcolor{Blue}{\textbf{\textsc{JAC}}} is a man and never switches languages. \textcolor{violet}{\textbf{\textsc{ASH}}} speaks first.  \\
\hline

\end{tabularx}
\caption{Examples of prompts for two speakers ID'd \textcolor{violet}{\textbf{\textsc{ash}}} and \textcolor{Blue}{\textbf{\textsc{jac}}}, structured in the three different forms: List, Sentence, and Partner. We prepend these prompts to dialogue context $D$ to train our speaker-grounded models. All prompts cover attribute set $\mathcal{A}$ consisting of age, gender, country of origin, language preference, code-switching preference, and speaker order in the global dialogue context. Sentence and List prompts are similar in that they describe speakers separately; Sentence prompts are more prose-like. Partner prompts first highlight \textcolor{ForestGreen}{similarities} between speakers, capturing speaker entrainment features, before describing unique features of each speaker.}
\label{tab:descriptions}
\end{table*}

\label{sec:methodology}
In this section we introduce the task of predicting code-switching points and describe the base model for it, with a self-explainable architecture as its backbone. We then describe how we incorporate speaker-grounding prompts into the model.

\subsection{Task Definition}
\label{sec:meth-task}
Let $d_i = [w_1, w_2, \ldots, w_u]$ be an utterance (string of tokens) in the full dialogue $\mathcal{D}$. Given a context window of size $h$, a model processes a local dialogue context:
$[d_{i-h}, \ldots, d_{i-1}, d'_i]$, where $d'_i \coloneqq [w_1, w_2, \ldots, w_b], \* b \in \left\{1, 2, \ldots, u \right\}$. In other words, we take the prefix of the current utterance $d_i$ up to an index $b$. Each word $w_j$ in the dialogue has a language tag $l_j$ associated with it. 
 For the given dialogue context $D$ up to boundary-word $w_b$, a model must predict
 whether the language of the next word after $w_b$ will be \textit{code-switched} (1), or the same (0). In our setup, a code-switch occurs between two consecutive words $w_b, w_{b+1}$ if the language of $w_b$ is English and the language of $w_{b+1}$ is Spanish (or vice versa). In particular, a word with an ambiguous language, such as the proper noun \textit{Maria}, cannot be a switch point; only words with unambiguous language tags are switched. This prevents us from labeling monolingual utterances as code-switched only because they have an ambiguous term such as a proper noun.

\paragraph{Speaker-Aware Grounding}
\label{sec:meth-sp-def}
Each utterance in the dialogue context has a speaker associated with it. Let the set of all speakers in the dialogue context be $\mathcal{S} = \left\{ s_1, s_2, s_3, \ldots, s_M \right \}$. We define a speaker-aware prompt $\mathcal{P} =  \left\{ p_1, p_2, p_3, \ldots, p_K \right \}$ as a concatenation of $K$ strings $p_i$, each describing an attribute of a speaker in the dialogue. Together, $\mathcal{P}$ describes the unique attributes of all $M$ speakers in the dialogue context.

Our proposed speaker-guided models take as input $\mathcal{P} \cdot \mathcal{D} = [p_1, \ldots, p_K, d_{i-w}, \ldots, d'_i]$, the concatenation of prompts and dialogue context. We encode the inputs with a multilingual Transformer-based architecture \cite{bert-19, conneau-etal-2020-unsupervised} before using a linear layer to predict the presence or absence of a code-switch. 

\subsection{Generating Speaker Prompts}
\label{sec:meth-sp-gen}

We incorporate global information about each speaker in a dialogue using different prompt styles, generating a prompt $\mathcal{P}$ for a given dialogue context $D$. In theory, these prompts have the potential to change the model's priors by contextualizing dialogue with speaker information and should be more useful for predicting upcoming language switches. We consider two aspects when designing prompts.  

\paragraph{Content}  The prompt describes all speakers $\mathcal{S}$ in the dialogue using a set of speaker attributes $\mathcal{A} = \left\{a_1, a_2, \ldots, a_T \right\}$. To create a description $P_m$ for speaker  $s_m \in \mathcal{S}$, we combine phrases $p_{s_{m_{1}}}, p_{s_{m_{2}}}, \ldots, p_{s_{m_{T}}}$, such that each phrase corresponds to exactly one attribute. As Table \ref{tab:descriptions} indicates, we use speaker IDs to tie a speaker to her description, and all prompts cover the full set of attributes, $\mathcal{A}$, for all speakers in $D$.


 \paragraph{Form} We consider three prompt forms: \textit{List}, \textit{Sentence}, and \textit{Partner}. The prompt form determines both the resulting structure of prompt string $\mathcal{P}$ and the way we combine local attribute phrases $p_j$ to generate a speaker description $P_i$. Table \ref{tab:descriptions} provides concrete examples of List, Sentence, and Partner prompts for a pair of speakers.


 List and Sentence prompts do not explicitly relate speakers to each other: the final prompt $\mathcal{P} = \left\{P_1, \ldots, P_m, \ldots, P_M \right\}$ concatenates individual speaker prompts $P_i$. List forms combine all attributes in a speaker description $P_m$ with commas, while Sentence forms are more prose-like. These prompt forms are most straightforward to implement and simply concatenate each speaker profile without considering interactions of features. The model must implicitly learn how attributes between different speakers relate to one another in a way that influences code-switching behavior. 

\label{sec:methods_partner} 
Speaker \emph{entrainment} or accommodation influences code-switching behavior \cite{10.1145/3392846, ahn2019codeswitch, myslin-levy2015,parekh2020understanding}. Thus, we also created Partner prompts to explicitly highlight relationships between speakers. We hypothesize that these are more useful than the List and Sentence forms, from which the model must implicitly learn speaker relationships. Partner prompts include an initial $P_i$ containing attribute qualities that all speakers share:
$$P_i \coloneqq \left\{p_{a_j} | a_j = v_k,\*  \forall s \in \mathcal{S} \right\},  $$
where $a_j \in \mathcal{A}$ and $v_k$ is a value taken on by attribute $a_j$. As an example, all speakers may prefer Spanish, so $P_i$ will contain an attribute string $p_i$ capturing this. The final partner prompt is $\mathcal{P_{\textit{partner}}} = \left\{P_{i}, P_1, P_2, \ldots, P_M\right\}$,  where speaker-specific descriptions $P_1, P_2, \ldots, P_M$ highlight unique values of each speaker. 

We prepend prompts $\mathcal{P}$ to dialogue context $\mathcal{D}$ using \textsc{[eos]} tokens for separation. We do not vary the feature order in a given prompt, but additional prompt tuning may reveal an optimal presentation of features in these prompts.


\subsection{Interpretable Text Classification}
\label{sec:meth-selfexp}
Our proposed setup takes as input the dialogue context and a prepended speaker prompt. To explain predictions of the baseline and our speaker-aware setups, we use SelfExplain \cite{rajagopal2021selfexplain}, a framework for interpreting text-based deep learning classifiers using phrases from the input. SelfExplain incorporates a Locally Interpretable Layer (LIL) and  a Globally Interpretable Layer (GIL). GIL retrieves the top-k relevant phrases in the training set for the given instance, while LIL ranks local phrases within the input according to their influence on the final prediction. LIL quantifies the effects that subtracting a local phrase representation from the full sentence have on the resulting prediction. We exclusively use LIL to highlight phrases in the speaker prompts and dialogues to identify both social factors and linguistic context influential to models; through post-hoc analysis, we can reveal whether these features can be corroborated with prior literature or indicate a  model's reliance on spurious confounds. We do not use the GIL layer because we do not have instance-level speaker metadata; instead, speaker features are on the dialogue-level and will not yield useful top-k results. Figure \ref{fig:architecture} illustrates our full proposed model with two classification heads: one for prediction and one for interpretation. \Sref{sec:phrase-scores} describes how we score phrases according to their influence on the final prediction. 

\section{Ethical Considerations}
\label{sec:ethics}
\paragraph{Data Privacy} In line with prior behavioral studies, our work illustrates that sociolinguistic cues are essential for predicting code-switching points. To deploy our speaker-informed model, we must protect the identity and privacy of users through techniques such as federated machine learning: deploying local models to end-users without sending any user information back to the cloud \cite{konecny}. Local models and data should be encrypted to prevent breaches and tampering with algorithms, as well as possible reconstruction of training data \cite{hitaj2017deep, carlini2019secret, zhang2020secret}, minimizing the risk of leaking speaker information. 
Additionally, deployed systems should only collect and access information if the user agrees to it.  All conversational participants voluntarily shared the data we use. 

Moreover, this research is important to conduct because
there is evidence that human users react positively to appropriately adaptive technologies \citep{branigan2010linguistic}. Specifically, initial experiments indicate that users rate dialogue systems that incorporate code-switching higher than ones that do not (or that do it less naturally)  \citep{ahn2019codeswitch,10.1145/3392846}. A classifier, such as the one we explore in this work, can be very useful for developing a naturalistic dialogue system that is useful and enjoyable to use by people of diverse linguistic backgrounds. Our work focuses on English-Spanish code-switching which is widespread and accepted, but different regions and cultures have varying opinions of code-switching. It is important to understand these before building an application for a new language pair \cite{dogruoz-etal-2021-survey}.
 \section{Experimental Setup}
 \label{sec:experimental}
\subsection{Dialogue Data}
\label{sec:data}
Our task requires a dataset which not only has natural, mixed-language dialogue, but includes also information about its speakers. We use the Bangor Miami \cite{bangor} dataset (BM) containing 56 transcribed dialogues in mixed English and Spanish. Most dialogues are between two speakers, but may contain three or four; another set of dialogues records only one speaker's side of the conversation. These ``monologues'' are still useful to study how linguistic cues influence code-switching. Moreover, language IDs are provided for every token. The dataset includes a questionnaire of self-reported information about each conversational participant; this includes dialogue-independent, macro-social features such as age, gender, and country of origin, as well as language preferences and speaker-provided linguistic ability. We identify each country according to the primary language (English, Spanish, or neither) spoken in the country and bin age features into four comparative groups ranging from youngest to oldest.  An order feature indicates which speaker spoke first, second, etc.\ in the global dialogue context; we hypothesize that speakers may entrain, or change their speech to match, those who start a conversation \cite{ahn2019codeswitch}. Altogether, six features define our attribute set $\mathcal{A}$.
\subsection{Code-switching Dataset Creation}
 
 For each dialogue in BM, we extract all existing code-switch points; for a given switched word, we retain all left-most context in its containing utterance and vary the number of prior utterances that are included as context between 1, 2, 3, and 5. 
 To generate negative examples, we select monolingual utterances by sampling from a binomial distribution with $p=0.75$. For each retained utterance, we randomly choose three potential switch points (extracting leftmost context in the same way), resulting in a dataset that is approximately 25\% switched. 

\paragraph{Creating Splits}
Most speakers participate in only one of the 56 dialogues in the corpus. To help ensure the model sees new dialogue context in training and testing time,  we split the train, validation, and test splits by conversation in a 60:20:20 ratio. For each dialogue, we compute the multilinguality index (M-Index) \cite{barnett2000lides}, a measure between 0 and 1 indicating the mixedness in the text: 0 is monolingual text, while 1 is a code-switch at every word. We stratify the conversations by the M-Index and code-switching labels to enforce a more balanced distribution of monolingual and mixed-language conversations.
\begin{table}[b!]
\centering
\begin{tabular}{lrrl}
\hline \textbf{Set} & \textbf{Validation} & \textbf{Test} \\ \hline
Balanced & 0.500 & 0.500\\
Unbalanced & 0.250 & 0.252\\

\hline
\end{tabular}
\caption{\label{tab:cs-prop-test-stats} Proportion of code-switched examples in the balanced and unbalanced validation and test splits. }
\end{table}

We down-sample monolingual examples to balance training and validation splits and report results on unbalanced validation and test sets. Table \ref{tab:cs-prop-test-stats} shows the proportions of code-switched examples. Our final balanced training and validation splits have about 14,000 and 3,000 examples, while the unbalanced validation and test sets have approximately 7,000 and 9,000 examples, respectively.

\paragraph{Marking Dialogue Turns}
 The baseline setup does not incorporate speaker cues. Instead we use \textsc{[eot]} and \textsc{[eou]} tokens at the end of each utterance to signify end-of-turn and end-of-utterance, respectively. Given two consecutive utterances, an \textsc{[eot]} signifies a change in speakers, while  \textsc{[eou]} indicates no change. In the speaker-informed setup, unique speaker IDs distinguish utterances from each speaker, and we prepend informative \textit{prompts} characterizing the conversational participant(s). Prompts include user-reported metadata of personal preferences and characteristics. We use three prompt templates, as detailed in Section \ref{sec:methodology}.

\subsection{Training Details}
We use XLM-RoBERTa (XLMR) \cite{conneau-etal-2020-unsupervised} to encode the text and jointly fine-tune XLMR on the code-switch prediction task. As a baseline, we use an XLMR model without prompt inputs $\mathcal{P}$. Our speaker-prompted models, SP-XLMR, are trained by prepending speaker prompts to the dialogue context. The small size of our dataset results in higher variability in performance. To mitigate this, we select 10 random seeds, and train a given model setup (i.e., list prompt, no prompt, etc.) on each seed. The number of seeds is arbitrary; however, we choose a generous number of seeds to yield a tighter confidence interval for our results. We use 3 prompt types, resulting in 30 speaker-prompted models and 10 baseline models. We refer to speaker-prompted models as \textsc{sp-xlmr} and to the non-speaker baseline as simply \textsc{xlmr}. All models are trained using AdamW optimizers with a weight decay of $1e^{-3}$ for a maximum of 10 epochs. \textsc{sp-xlmr} models are trained with a learning rate of $5e^{-5}$ and \textsc{xlmr} models use a learning rate of $1e^{-5}$. To refer to a particular speaker-prompted model, we use a combination of prompt form and context size, for example, \textsc{list-5}.

We report accuracy, F1, precision, and recall on the unbalanced validation and test sets. We use the Mann Whitney U significance test because it does not assume normally-distributed population means.


\section{Evaluation}
\label{sec:eval}
\paragraph{Speaker prompts significantly improve code-switch prediction.}
Table \ref{tab:val-by-prompt} includes average accuracy and F1 of \textsc{xlmr}, \textsc{list}, \textsc{sentence}, and \textsc{partner} models, across all context windows and seeds on the unbalanced validation and test sets. Each value is an average of 40 models. Adding prompt features boosts accuracy upwards of~5-8 percent points and F1 by~.04-.05  compared to \textsc{xlmr};  \textsc{xlmr} does not even surpass the majority baseline in accuracy. Based on validation set results, partner features are most helpful, confirming our sociolinguistically-driven hypothesis (see Section \ref{sec:methods_partner})  Moreover, the standard deviation of \textsc{xlmr} accuracy is more than twice as large (3.66 on validation and 2.95 on test) as that of any speaker-prompted model. The improvements in accuracy and decrease in variation between models suggest that explicit speaker information guides models to learn relevant inductive biases for the code-switching task. However, we cannot guarantee that the trained models will not reveal harmful social biases in other tasks. 

We see similar trends, regarding accuracy, F1, and standard deviation, in Table \ref{tab:val-by-ctx}, which includes results for \textsc{sp-xlmr} and \textsc{xlmr} across the different context windows; each \textsc{sp-xlmr} and \textsc{xlmr} value is an average of 30 and 10 models, respectively. Larger context windows are helpful for both model types. Tables \ref{app:val_table} and \ref{app:test_table} include precision and recall scores for each prompt type and context window; in general, speaker-prompted models have upwards of .10 points higher precision than baseline \textsc{xlmr}, indicating that speaker information helps to identify valid switch points. As the context window increases, all speaker prompt types yield fairly similar performance. However, when context sizes are small (1 or 2 previous utterances only), Partner and Sentence prompts yield higher accuracy and precision than List models, perhaps because these prose-like formats are more useful for the model than a simple concatenate list of features.

\paragraph{Using irrelevant speaker descriptions worsens model performance.} As a control, we generated synthetic descriptions for each speaker, including features such as favorite foods and weather, owned pets, and height. None of these attributes are discussed in the conversations and would not explicitly influence code-switch production. After generating descriptions in the Sentence and Partner format, we prepend them to dialogues using a context window of 5. According to the results in Table \ref{tab:speaker-control}, these pseudo-descriptions significantly decrease performance, even relative to the baseline \textsc{xlmr} model trained with a context window of 5. The results indicate that domain knowledge is useful to understand which speaker features to add to a model to improve performance, and they give more support to the claim that \emph{relevant} speaker information helps guide models to useful inductive biases. 

\begin{table}[t!]
\centering
\scalebox{0.9}{
\begin{tabular}{c|cc|cc}
\multicolumn{1}{c|}{}&\multicolumn{2}{c|}{\textbf{Validation}} & \multicolumn{2}{c}{\textbf{Test}}\\

\hline \textbf{Model} & \textbf{Acc. (\%)} & \textbf{F1} & \textbf{Acc. (\%)} & \textbf{F1}\\ \hline
Majority & 75.0  & -- & 74.8 & --\\
Minority & 25.0 & .294 & 25.2 & .296 \\\hline
XLMR & 70.3 $\pm${3.66} & .573 & 72.0 $\pm${2.95} & .591 \\
    List & 77.6 $\pm${1.68} & .615 & \textbf{79.7} $\pm${1.13} & \textbf{.632} \\
Sentence & 78.1  $\pm${1.60}& .618 & 79.5  $\pm${1.31}& .630\\
Partner & \textbf{78.3}  $\pm${1.58} & \textbf{.621} & 79.4  $\pm${1.50} & .622 \\

\hline
\end{tabular}
}
\caption{\label{tab:val-by-prompt} Average accuracy and F1 scores of prompt models and  XLM-R on validation and test sets. There are $N$$=$$40$ models for all setups. Majority and Minority baselines are included for comparison. \textbf{Bold} scores indicate the best performance on the split. All results are significant ($p<0.0001$) by Mann-Whitney U Tests.}

\end{table}

\begin{table*}[t!]
\centering
\begin{tabular}{l|cc|cc|cc|cc}
\hline & \multicolumn{4}{c|}{\textbf{Validation}} & \multicolumn{4}{c}{\textbf{Test}}\\
\hline  & \multicolumn{2}{c}{\textbf{SP-XLMR}} & \multicolumn{2}{c|}{\textbf{XLM-R}}
& \multicolumn{2}{c}{\textbf{SP-XLMR}} & \multicolumn{2}{c}{\textbf{XLM-R}}
\\ \hline
\textbf{Ctx} & \textbf{Acc. (\%)}  &\textbf{F1} & \textbf{Acc. (\%)}  & \textbf{F1} & \textbf{Acc. (\%)}  & \textbf{F1} & \textbf{Acc. (\%)}  & \textbf{F1} \\ \hline
1 & 76.9 $\pm${1.96}  & .605  & 66.4 $\pm${2.84} & .540    & 78.8 $\pm${1.54}  & .607  & 69.5 $\pm${2.75} & .565  \\
2 & 77.9 $\pm${1.10} & .615  &  70.3 $\pm${3.27}  &  .572 & 79.6 $\pm${1.13}  & .629 &  71.8 $\pm${2.23} &  .587 \\
3 & 78.6 $\pm${1.17} & .622  & 71.4 $\pm${1.92}  & .582   & 80.0 $\pm${0.96} & \textbf{.636}  &  72.4 $\pm${2.31} & .598\\
5 & 78.7  $\pm${1.56} & \textbf{.631}  & 73.1 $\pm${2.74}  & .598  & 79.7 $\pm${1.34}  & \textbf{.639}  &  74.2 $\pm${2.39} & .612 \\

\hline
\end{tabular}
\caption{\label{tab:val-by-ctx} Average Accuracy and F1 of prompt models and baseline XLM-R on validation and test sets, for $N$$=$$30$ SP-XLMR models and $N$$=$$10$ XLMR models. All results are significant ($p<0.0001$) by Mann Whitney U Tests.}

\end{table*}

\section{Explaining Performance Gaps}
\label{sec:analysis}
Compared to baseline models, speaker models not only attain higher accuracy and F1 scores, but they also have a much smaller standard deviation in scores. For these experiments, we seek to explain our findings using the important phrases identified by LIL.  Within a speaker prompt $\mathcal{P}$, each speaker characteristic maps to its own phrase (i.e., \textit{from an English-speaking country}); in the dialogue, we extract 5-gram phrases using a sliding window. We detail our approach to scoring phrase influence and analyze key dialogue and speaker features.
\subsection{Computing Phrase Relevance}
\label{sec:phrase-scores} 
Our goals are to (a) identify phrases in the input whose removal will change the resulting model prediction and (b) identify phrases which contribute high confidence to the resulting model prediction. Let $F$ be the full textual input consisting of sole dialogue context or dialogue context prepended with prompts, while $Z_{F}$ is the softmax output from our classifier. Let $j$ be the index of the class predicted from the full input. LIL inputs $Z_{F}$ along with a series of masks, each corresponding to a local phrase in either the dialogue or the speaker prompt. Let $\textit{nt}$ be a local phrase, such that $nt$ is either a speaker phrase $p_i$ or an n-gram in an utterance $d_i \in D$. Using LIL, we quantify the effect of removing the representation of phrase \textit{nt} from the representation of $F$ by comparing the activation differences of   $Z_{nt}$ and $Z_f$ at index $j$, and we analyze the resulting sign and magnitude to address goals (a) and (b), respectively:  
\begin{align}
\label{eq:se_sign}
                  C \coloneqq \left\{ \begin{array}{ll}
                    1 & \arg\!\max Z_{\textit{nt}} = j \\
                                -1 & \arg\!\max Z_{\textit{nt}} \neq j
                        \end{array} 
                        \right.       
\end{align}
\begin{align}
\label{eq:se_score}
            & \textit{r}({\textit{nt}})  = C \: |z_{{\textit{nt}}_{j}} - z_{{F}_{j}}| 
    \end{align}
\noindent
where $z_{{nt}_{j}}$ and $z_{{F}_{j}}$ are the softmax scores of the phrase-ablated sentence and the full sentence, respectively, at index $j$, and $r(nt)$ is the relevance score of $nt$.
As Equations \ref{eq:se_sign} and \ref{eq:se_score} indicate, we analyze a local phrase's score as follows: 
\begin{compactdesc}
    \item[Sign] A positive sign ($C = 1$) indicates that the representation without \textit{nt} does not change the resulting prediction. A negative score ($C=-1$) indicates a more influential phrase because its ablation results in a different prediction. 
    \item[Magnitude]  corresponds to the weight of the contribution of a particular phrase. If the activation difference is high in magnitude, then \textit{nt} strongly influences the resulting prediction. Magnitudes near 0 indicate a non-influential phrase.
\end{compactdesc}

Our scoring approach differs slightly from the original implementation (see Appendix \ref{app:score-details}).
\subsection{Analyzing Dialogue Phrases}

 Given a context size, the dialogue phrase masks are identical for SP-XLMR and XLMR; thus, we directly compare which phrases are most informative in the presence and absence of speaker features. We consider only phrases which are influential enough to change a given model's prediction after their representations are subtracted from the full-sentence representations (phrases with a negative score). 

Setting context size to 5, we identify examples from the validation set for which the majority of SP-XLMR models (out of 30) predicted correctly and the majority of XLMR models (out of 10) predicted incorrectly. Nearly 95\% of such examples are not switched, indicating that added speaker information helps improve model precision. We sample a portion of these instances for our analysis.

For a given validation set example and model setup, we track all influential phrases and count the number of models for which each phrase is influential. To account for phrase interactions, we track the agreement on co-occurring pairs and trios of important phrases. We compare only top-10 influential phrases. We use 10 phrases because all models rank at least 10 phrases as influential (but not 15 or 20). Phrase scores in the top-$k$, where $k<10$, tend to all be very similar. We are not interested in small-scale score differences, and thus, equally consider all phrases ranked in the top-10. We hypothesize that speaker models (1) exhibit more phrase agreements compared to baseline models and (2) use more helpful and relevant linguistic features for code-switch prediction. 

\begin{figure}[htb]
    \centering
    \includegraphics[scale=0.5, clip, trim=2.7cm 15cm 2.6cm 3cm]{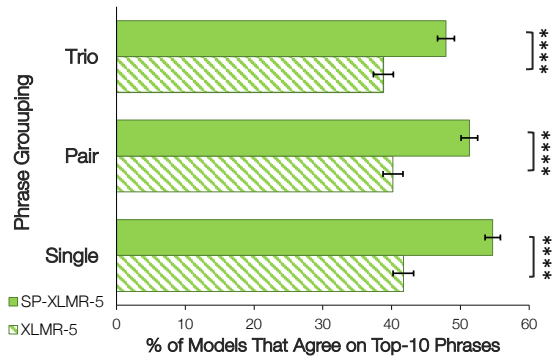}
    \caption{Each bar indicates the average percent and standard deviation of XLMR (dashed green, $N$$=$$10$) and SP-XLMR (green, $N$$=$$30$) models that agree on the top-10 phrases. We consider single phrases, as well as pairs and trios of phrases. There is significantly less agreement ($p<0.0001$) among XLMR models as compared to SP-XLMR, potentially accounting for the higher standard deviation in XLMR models' scores.}
    \label{fig:inter-model-agreement-spk}
\end{figure}

\paragraph{Most speaker models agree on which phrases are important.} In addition to tracking which individual phrases are in the top-10, we analyze how many pairs and trios of phrases are in the top-10 list. Figure \ref{fig:inter-model-agreement-spk} indicates that the majority of speaker-prompted models (out of 30) tend to agree on the top-10 important phrase groupings, especially across single and pairwise groupings. The speaker models likely pick up on similar inductive biases, as revealed through the higher feature agreement among these models.  Only around 38-40\% of baseline models tend to agree on which phrases are most important, potentially explaining the higher standard deviation in results among the baseline models compared to the speaker models. 

\paragraph{Speaker models make better use of language information.} On monolingual (negative) examples, both speaker-prompted and baseline models tend to look at a majority of monolingual phrases in the same languages (English or Spanish), and these phrases are mainly located in the first quarter of tokens preceding the potential switch point. However, speaker models successfully predict many of these negative examples correctly, unlike baselines. In many cases, the speaker models have additional access to global speaker properties of the current speaker -- for example, \textit{never switches languages} -- and this may also influence them to make the correct prediction given prior linguistic context. Even when baseline models have strong evidence for predicting no code-switch (i.e., ranking only monolingual phrases as important), they tend to misuse this history and randomly predict code-switches. 

On code-switched examples, speaker models continue to favor phrases that are nearest to the switchpoint, while baseline models are sensitive to phrases in early and late dialogue context. Using phrases closer to switch points may give better structural context from which to predict a switch. In several cases, speaker models correctly predict an English-to-Spanish switch and rank prior Spanish phrases as influential, while baseline models highly rank English phrases and predict no switch. We see a similar pattern in Spanish-to-English switches. Speaker information may help models learn, linguistically, what it means to code-switch.

\subsection{Analyzing Speaker Phrases}
\paragraph{Linguistic preference features are most influential across model setups.} For all speaker-prompted models, speakers' language and code-switching preferences are the most influential on the resulting predictions. Country of origin information is helpful, too, but may be misleading: speakers may immigrate from a Spanish country but grow up speaking English; in such cases, the language information likely helps disambiguate any confusions. Following these linguistic features are relational features (speaker order) in the dialogue, and less often,  age features, especially in partner models. Gender is almost never influential. Age may be correlated with linguistic preference features and is thus not influential on its own. Gender and age can interplay to define larger, dynamic social roles, which may influence language production. On their own, these static markers of identity do not significantly characterize one's speech patterns   \cite{eckert2012three, ochs199214}. However, shared macro-social attributes (age and gender) may be influential in partner models because the ``participant constellation'' influences how speakers express themselves and modulate social distance \cite{giles2008communication, myslin-levy2015}.

\paragraph{Linguistic preferences of the group are most influential in true and predicted code-switches.} To study how linguistic preferences interact with code-switch behavior, we analyzed all true and predicted code-switch points according to the preferences of speakers in the conversations. Specifically, we tracked whether speakers preferred to code-switch, speak English, speak Spanish, or preferred both English and Spanish. Looking at each feature in isolation, we counted the number of switch points that occur when at least one speaker prefers the feature, when the current speaker prefers the feature, and when other conversational partners (aside from the current speaker) prefer the feature. Table \ref{tab:pref} in the Appendix indicates that preference to code-switch, to speak Spanish, or to speak both English and Spanish are most dominant in influencing code-switching behavior. Speaker-prompted models are able to learn this relationship. Moreover, we found that preferences of speakers other than the current speaker tend to be more influential in driving code-switching behavior, relating to the idea of speaker entrainment or accommodation. For more details, please refer to Appendix \ref{app:speaker-ling-pref}.

\paragraph{Ablating Features} 
Using the best-performing setups on the validation set, namely Partner and Sentence models with 5 prior utterances for context, we identify influential speaker attributes using a leave-one-out-approach to mask out each attribute $a_i \in \mathcal{A}$. For each attribute, we train 10 ablated models and evaluate on the validation set. Note that this is different from the phrase ablations using LIL because we finetune the XLM-R encoder during the training process; in this setup, the ablated feature information is never backpropagated to update the encoder weights. 

The results of these experiments (see Appendix \ref{app:speaker-ablation}) give some evidence that language preference, mixing, and age information have statistically significant effects on the performance of Partner-5 models, but this does not hold for the Sentence-5 models. We have strong evidence to believe that these speaker attributes have more complex underlying relationships and leave the exploration of these multi-feature interactions for future work. 


\section{Related Work}
\label{sec:related}
Our use of prompts\footnote{Our prompts are data-dependent and fixed, and thus rather unrelated to the prompt tuning literature \citep{liu2021pre}.} is similar to \citet{zhong2021adapting} and \citet{wei2021finetuned}, who rely on prompts to put models in different states for different tasks. 

\paragraph{Speaker Personas} Open-domain dialogue agents which act according to a persona are more natural and engaging than the non-personalized baselines \cite{li-etal-2016-persona}; these personas can be short, superficial descriptions generated through crowdsourcing \cite{zhang2018personalizing}, gathered from Reddit \cite{mazare-etal-2018-training}, or self-learned (inferred) from dialogue context  \cite{madotto-etal-2019-personalizing, cheng-etal-2019-dynamic}. These works, however, primarily evaluate dialogue \textit{content} and only in one language (English) instead of analyzing how speaker properties influence the downstream dialogue structure. 

\paragraph{Addressing Model Bias} Prior works for mitigating social biases feature adversarial learning \cite{pryzant2018deconfounded,elazar2018adversarial}, 
counterfactual data augmentation    \citep{zmigrod-etal-2019-counterfactual,Kaushik2020LearningTD} or dataset balancing \cite{zhao2017men}, and more recently, using an interpretability-driven approach to uncover and controllably demote hidden biases \cite{han21}. Techniques for adapting to linguistic variants and mixed-language data include adversarial learning to pick up on key linguistic cues \cite{kumar21}, augmenting datasets with synthetic text \cite{winata-syth-19} or examples of variants that models underperform on \cite{chopra2021switch}, discriminative learning \cite{gonen-yoav-18-discrim}, and transfer learning with morphological cues \cite{aguilar-transfer-19}. 


\paragraph{Codeswitch Prediction} The first work in code-switch prediction \cite{solorio2008learning} uses Naive Bayes (NB) on lexical and syntactic features of shallow word context before switch boundaries from a small, self-collected dataset of English-Spanish conversations. Another NB approach predicts switch points on Turkish-Dutch social media data \cite{papalexakis2014predicting}, additionally using multi-word expressions and emoticons in their experiments. \citet{Piergallini2016WordLevelLI} extend the techniques of the prior two works to Swahili-English codeswitched data. Two fine-grained logistic regression analyses \cite{fricke2016primed, myslin-levy2015} go beyond lexical information, incorporating psycholinguistic properties such as word accessibility and priming effects, and include binary features to code for properties such as speaker age and preceding utterance language.

\section{Conclusion}
\label{sec:disc-conc}

To the best of our knowledge, this is the first work incorporating sociolinguistically-grounded social factors in an interpretable neural model for code-switch point prediction. %
Our speaker-aware models can better leverage mixed-language linguistic cues, compared to a text-only baseline: specifically,
we showed performance gains of up to 7\% in accuracy and .05 points in F1 scores on an imbalanced code-switching dataset. Our work is limited to one language pair and uses a small dataset. Thus, additional studies are necessary to assess the generalizability of our findings to other languages. Moreover, speaker identities can change dynamically in different settings. Linguistic preferences may also change over time. We could move beyond static personas, refining them using local dialogue context. In addition, speaker-grounded models must be carefully engineered to protect user privacy, using proxies for personal information and keeping private information away from shared resources. 

In the future, we would like to explore whether such speaker prompting can improve models in \emph{other} person-centered tasks, e.g., coreference resolution (especially for datasets explicitly testing gender biases) or sentiment analysis. Using techniques such as data augmentation, we can explicitly guide models away from biases learned during training. With ethical considerations in mind, our work advances the state-of-the-art in  building more adaptable and person-aware NLP technologies.

\section*{Acknowledgements}

We thank Vidhisha Balachandran, Dheeraj Rajogopal, Xiaochuang Han, Artidoro Pagnoni, and the anonymous reviewers for providing valuable feedback on our work. This work was supported in part by grant No.~2019785 from the United States-Israel Binational Science Foundation (BSF), National Science Foundation (NSF) grants No.~2007960, 2007656, 2125201 and 2040926, and by grant No. LU 856/13-1 from the Deutsche Forschungsgemeinschaft (DFG).
\bibliography{anthology,custom}
\bibliographystyle{acl_natbib}
\

\renewcommand{\dbltopfraction}{0.9}	
\renewcommand{\textfraction}{0.07}	
\renewcommand{\floatpagefraction}{0.7}	
\renewcommand{\dblfloatpagefraction}{0.7}	

\appendix
\section{Appendix}
\subsection{Code-Switching Example}
\label{app:cs-example}
\begin{figure}[hbt]
    \centering
    \includegraphics[clip, trim={33cm 0cm 3.5cm 0cm}, scale=0.23]{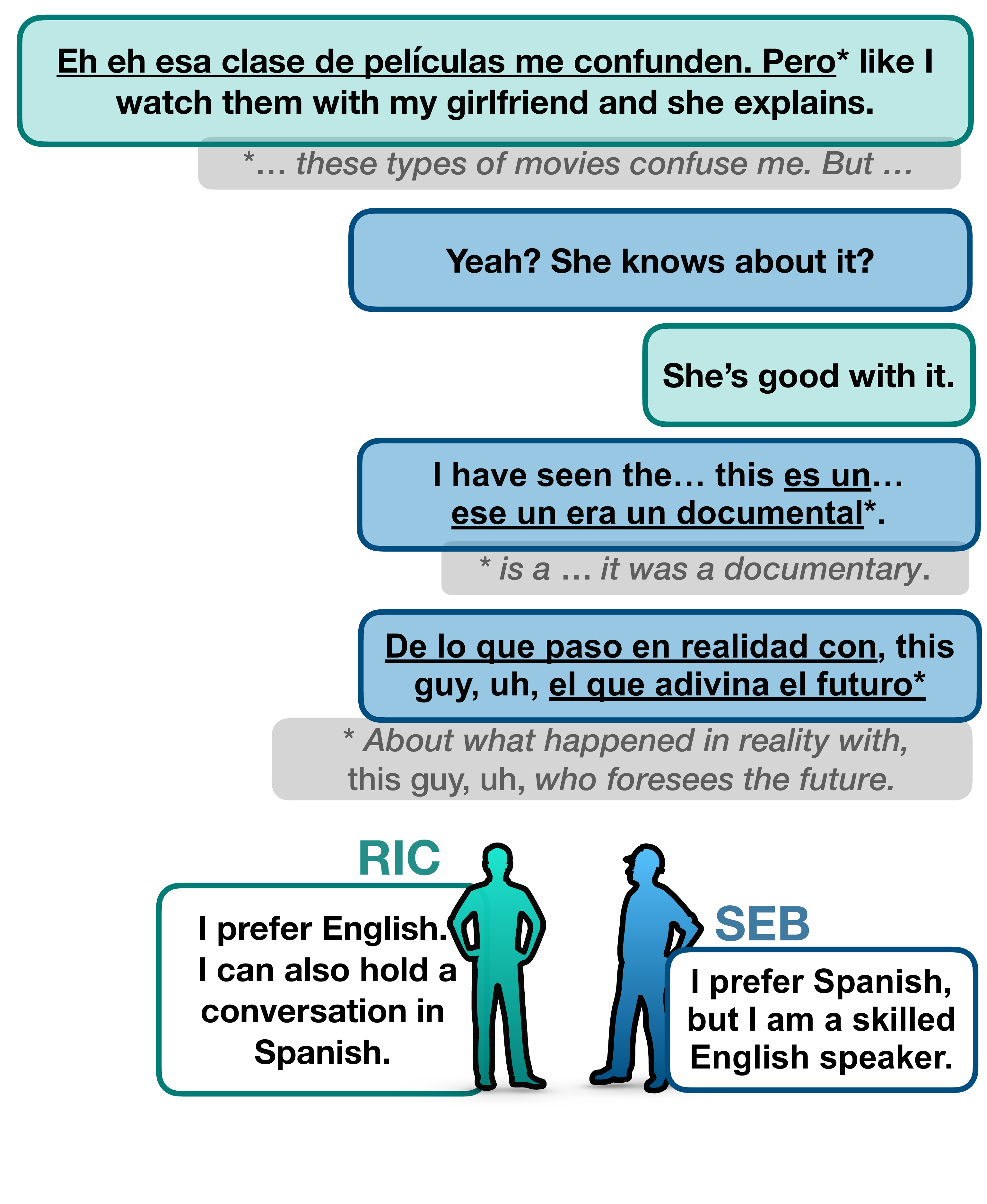}
    \caption{A dialogue between two speakers, whose IDs are \textsc{ric} and \textsc{seb}. \textsc{ric} and \textsc{seb} typically switch to to the languages they prefer: English and Spanish, respectively. \textsc{ric} and \textsc{seb} also mix languages to accommodate each other, demonstrating the need for speaker awareness in code-switched language processing.}
    \label{fig:cs_example_app}
\end{figure}

Figure \ref{fig:cs_example_app} provides a key motivating example of how global speaker features of two conversational participants, ID'd \textsc{ric} and \textsc{seb},  influence their local speech production. \textsc{ric} was raised in the United States and knows Spanish, while \textsc{seb} is from a Spanish-speaking country and has a strong grasp of English. For most of the dialogue,  \textsc{ric} speaks English, unless he is specifically accommodating to \textsc{seb}, as we see in the very first example utterance.   \textsc{ric} demonstrates more intrasentential (within-utterance) switches, often switching back to English, which corresponds to his preference for English \cite{cs-martinez}. \textsc{seb} accommodates to  \textsc{ric} by responding in English with \textit{Yeah, she knows about it?}, but, similarly to \textsc{ric}, relies on Spanish to express vocabulary or phrases that are more complex for him (i.e., \textit{foreseeing the future}).

\begin{table*}[hbt]
\centering
\scalebox{0.9}{
\begin{tabular}{l|c|c|c|c}
\hline & \multicolumn{2}{c|}{\textbf{Validation}} & \multicolumn{2}{c}{\textbf{Test}}\\
& \textbf{Acc. (\%)}  &\textbf{F1} & \textbf{Acc. (\%)}  & \textbf{F1}  \\ \hline
C-Partner-5 & 58.0 $\pm${15.3}  & .502     & 62.1 $\pm${17.1}  & .491  \\
C-Sentence-5 & 56.2 $\pm${16.9} & .372 & 61.4 $\pm${18.6} & .351  \\

\hline
XLMR-5 & 73.1 $\pm${2.74}  & .598 & 74.2 $\pm${2.39} & .612   \\
Partner-5  & 78.8 $\pm${1.22}  & .629  & 79.5 $\pm${1.27}  & .629 \\
Sentence-5  & 78.2 $\pm${1.86}  &  .632  & 79.4 $\pm${1.60}  & .644  \\
\hline
\end{tabular}
}
\caption{\label{tab:speaker-control} Average accuracy and F1 of Speaker Control models in the Partner and Sentence formats, with context size 5 (C-Partner-5 and C-Sentence-5, respectively). The descriptions contain synthetic speaker features that are not relevant to code-switching. We include the baseline XLMR-5, Partner-5, and Sentence-5 models for reference.}

\end{table*}

\subsection{Scoring Phrase Relevance}

\label{app:score-details}
Note that the original implementation scored phrases using the raw softmax difference to reflect the contribution of each phrase. Additionally, we use the sign of a score to indicate whether there is a change in model prediction. Consider case (1): the softmax score of the predicted class $z_{F} = 0.6$ and the phrase-ablated sentence yields a softmax score $z_{A} = 0.4$, while in case (2) the $z_{F'} = 0.9$ and $z_{A'} = 0.7$. Subtraction yields $z_{F'}-z_{A'}=0.2$, thus, the purported relevance of the ablated phrases is the same. However, we would score case (1) with -0.2 and case (2) with 0.2 to distinguish that in case (1), the ablated phrase changes the final prediction. 

\subsection{Speaker Control Results}
\label{app:speaker-control}
We synthesize a persona for each speaker describing preferred foods and weather, height (tall or short), and what kind of pet animal they have. These features are not relevant to code-switch production and are not discussed in any of the dialogues. We generate descriptions using the Sentence and Partner templates, set context size to 5, and following our methodology, train 10 models for each setup. Table \ref{tab:speaker-control} illustrates the results of these models on the validation and test sets. 

\subsection{Code-Switch Point Analysis }
\label{app:speaker-ling-pref}

We analyze the predicted and true code-switch points according to the linguistic preferences of speakers in the dialogue. Specifically, we consider four features: preference to code-switch, preference to speak English, preference to speak Spanish, and equal preference for both languages. For a given feature, we analyze each code-switch according to  whether (1) at least one conversational participant prefers this feature, (2) the current speaker prefers this feature or (3) at least one person, apart from the current speaker, prefers this feature. Table \ref{tab:pref} illustrates the percentage of code-switched points that occur when  conversational participants prefer a given feature, both in the dataset and in the predictions from the (P)artner, (S)entence, and (L)ist models that use a context size of 5.

Table \ref{tab:pref} illustrates that the Partner, Sentence, and List-5 models capture fairly well the relationship between speaker preference and the presence of code-switching. Specifically, the true percentages of code-switches that occur when conversational participants prefer to code-switch, or to speak English, Spanish, or both languages, are very close to those predicted by the three models.  Most code-switches occur when there is a speaker present who prefers to code-switch, to speak Spanish, or to speak both English and Spanish. The majority of code-switches (about 68\%) occur when at least one conversational participant prefers to speak both English and Spanish. More code-switches occur when at least one conversational participant, other than the current speaker, prefers a feature, indicating that speakers may accommodate their partners. This accommodation can influence communication more than the speaker's own preferences. 

\begin{table}[htb]
\centering
\scalebox{0.8}{
\begin{tabular}{c|c|c|c|c|c}
\multicolumn{1}{c|}{\textbf{Pref.}}&\textbf{Speaker}&\multicolumn{1}{c|}{\textbf{P-5}} & \multicolumn{1}{c|}{\textbf{S-5}} & \multicolumn{1}{c|}{\textbf{L-5}}
 & \multicolumn{1}{c}{\textbf{True}}\\

\hline
\multirow{3}{4em}{Switch} & Any & 44.3 & 45.8 & 44.2 & 44.4  \\
                       & Current & 20.4 & 20.8 & 20.2 & 19.6 \\
                       & Non-Current & 24.0 & 25.0 & 24.0 & 24.8 \\
\hline
\multirow{3}{4em}{English}& Any & 21.7 & 22.9 & 24.0 & 22.4  \\
                       & Current & 0 & 0 & 0  & 0 \\
                       & Non-Current & 21.7 & 22.9 & 24.0  & 22.4 \\
\hline
\multirow{3}{4em}{Spanish} & Any & 53.7 & 55.8 & 54.4 & 56.8  \\
                       & Current & 23.6 & 24.1 & 23.5  & 25.8 \\
                       & Non-Current & 30.0 & 31.7 & 30.9  & 31.0 \\
\hline
\multirow{3}{4em}{Both} & Any & 68.0 & 67.4 & 68.6 & 68.0  \\
                       & Current & 31.1 & 32.3 & 32.0  & 33.7 \\
                       & Non-Current & 36.6 & 35.2 & 36.6  & 34.2 \\

\hline

\end{tabular}
}
\caption{\label{tab:pref} Percentages of true and predicted code-switches that occur given the presence of speakers with different linguistic preferences. We consider four linguistic attributes: preference to code-switch, as well as preferences for English, Spanish, or both. We perform our analysis using true and predicted switch-points from the (P)artner-5, (S)entence-5, and (L)ist-5 models on the validation set. $N$$=$$10$ for all setups.}
\end{table}

\subsection{Speaker Ablation Results}
\label{app:speaker-ablation}
To analyze which speaker features influence code-switch predictions, we ablate a phrase, corresponding to one of six speaker features (age, gender, country of origin, language and code-switching preferences, and speaker order). Table \ref{tab:ablation} indicates that linguistic preferences are most influential.  
\begin{table}[bht!]
\centering
\scalebox{0.88}{
\begin{tabular}{c|cc|cc}
\multicolumn{1}{c|}{}&\multicolumn{2}{c|}{\textbf{Partner-5}} & \multicolumn{2}{c}{\textbf{Sentence-5}}\\

\hline \textbf{Feature} & \textbf{Acc. (\%)} & \textbf{F1} & \textbf{Acc. (\%)} & \textbf{F1}\\ \hline
Full & 78.9 $\pm$ 1.23 & .629 & 78.2 $\pm$ 1.86 & .632 \\ \hline 
Language & *76.9 $\pm${1.82} & .627 & 77.7 $\pm${1.84} & .629\\
Mixing & *77.8 $\pm${1.34} & .632 & 78.9 $\pm${1.72} & .636 \\
Country & 79.0 $\pm${1.54} & .635 & 78.4 $\pm${1.60} & .632 \\
Order & 78.9 $\pm${1.75} & .631 & 78.4 $\pm${1.30} & .632 \\
Gender & 78.0  $\pm${1.90}& .630 & 77.6  $\pm${2.12}& .627\\
Age & *77.5  $\pm${1.09} & .626 & 79.1  $\pm${1.80} & .634 \\

\hline
\end{tabular}
}
\caption{\label{tab:ablation} Average accuracy and F1 scores of speaker-ablated Partner-5 and Sentence-5 models on the validation set. $N$$=$$10$ for both setups. Full (non-ablated) models are included for comparison. Starred results are significant ($p<0.05$) by Mann-Whitney U Tests.}

\end{table}
\subsection{Validation and Test Results}
Tables \ref{app:val_table} and \ref{app:test_table} illustrate the results of all \textsc{xlmr} and \textsc{sp-xlmr} models over different context sizes on the validation and test sets, respectively.

\begin{table*}[htb]
\centering
\scalebox{1.0}{
\begin{tabular}{llcccc}
\hline \textbf{Model Type} & \textbf{Context} & \textbf{Acc. (\%)}  & \textbf{F1} & \textbf{Recall} & \textbf{Precision}\\ \hline
\hline
Majority & - & 75.0 & -- & -- & -- \\ 
\hline
List & 1 & 75.9 $\pm$ 1.390 & .601 & .723 & .515 \\
List & 2 & 77.4 $\pm$ 0.932 &  .609  &  .704 & .538 \\
List & 3 & 78.3 $\pm$ 1.191 & .620 & .710 & .553\\
List & 5 & 78.9 $\pm$ 1.418 & .631 & .721 & .564 \\
Partner  & 1  & 77.4 $\pm$ 2.301 & .611 & .706 & .541\\
Partner  & 2 &  78.3 $\pm$ 0.966 & .618 & .704 & .552 \\
Partner & 3  & 78.7 $\pm$ 0.982 &  .626 & .712 & .559 \\
Partner  & 5   &  78.8 $\pm$ 1.228 & .629 & .717 & .562\\
Sentence & 1 & 77.5 $\pm$ 1.667 & .605 & .689 & .542\\
Sentence  & 2 &  77.9 $\pm$ 1.210 & .616 & .710 & .546\\
Sentence  & 3  & 78.2 $\pm$ 1.255 & .621 & .695 & .563\\
Sentence  & 5 &  78.2 $\pm$ 1.863 & .632 & .745 & .551 \\
XLMR  & 1 & 66.4 $\pm$ 2.836 & .540 & .789 & .413\\
XLMR  & 2 & 70.3 $\pm$ 3.269 & .572& .790 & .452\\
XLMR  & 3 & 71.4 $\pm$ 1.916 & .582 & .796 & .460\\
XLMR  & 5 & 73.1 $\pm$ 2.739 & .598 & .799 & .480\\
\hline

\end{tabular}
}

\caption{Performance of all models on the validation set (25.0\% code-switched). Each value is an average of $N$$=$$10$ models. }
\label{app:val_table}
\end{table*}

\begin{table*}[htb]
\centering
\scalebox{1.0}{
\begin{tabular}{llcccc}
\hline \textbf{Model Type} & \textbf{Context} & \textbf{Acc. (\%)}  & \textbf{F1} & \textbf{Recall} & \textbf{Precision}\\ \hline

\hline
Majority & - & 74.8 & -- & -- & -- \\ 
\hline
List & 1 & 78.9 $\pm$ 1.247 & .614 & .667 & .572 \\
List & 2 & 79.7 $\pm$ 1.063 &  .630 & .685 & .584 \\
List & 3 & 80.0 $\pm$ 0.769 & .640 & .704 & .588 \\
List & 5 & 80.2 $\pm$ 0.919  & .643 & .709 & .590 \\
Partner  & 1  & 78.5 $\pm$ 1.966 & .598 & .635 & .570 \\
Partner  & 2 &  79.4 $\pm$ 1.120 & .624 & .678 & .581 \\
Partner  & 3 &  80.0 $\pm$ 0.927 & .635 & .690 & .589 \\
Partner & 5  & 79.5 $\pm$ 1.268 &  .629 & .692 & .581 \\
Sentence  & 1   &  79.0 $\pm$ 1.260 & .609 & .659 & .575\\
Sentence & 2 & 79.6 $\pm$ 1.095 & .634 & .704 & .579\\

Sentence  & 3  & 80.0 $\pm$ 1.021 & .634 & .694 & .586\\
Sentence  & 5 &  79.4 $\pm$ 1.602 & .644 & .742 & .573 \\
XLMR  & 1 & 69.5 $\pm$ 2.755 & .565 & .787 & .443\\
XLMR  & 2 & 71.8 $\pm$ 2.231 & .587 & .797 & .467\\
XLMR  & 3 & 72.4 $\pm$ 2.312 & .598 & .813 & .474\\
XLMR  & 5 & 74.2 $\pm$ 2.394 & .612 & .809 & .495\\
\hline
\end{tabular}
}
\caption{Performance of all models on the test set (25.2\% code-switched). Each value is an average of $N$$=$$10$ models.}
\label{app:test_table}
\end{table*}

\subsection{Model Architecture Diagram}
\begin{figure*}[htpb]

    \includegraphics[clip, trim=0cm 10cm 0cm 0cm, scale=0.23]{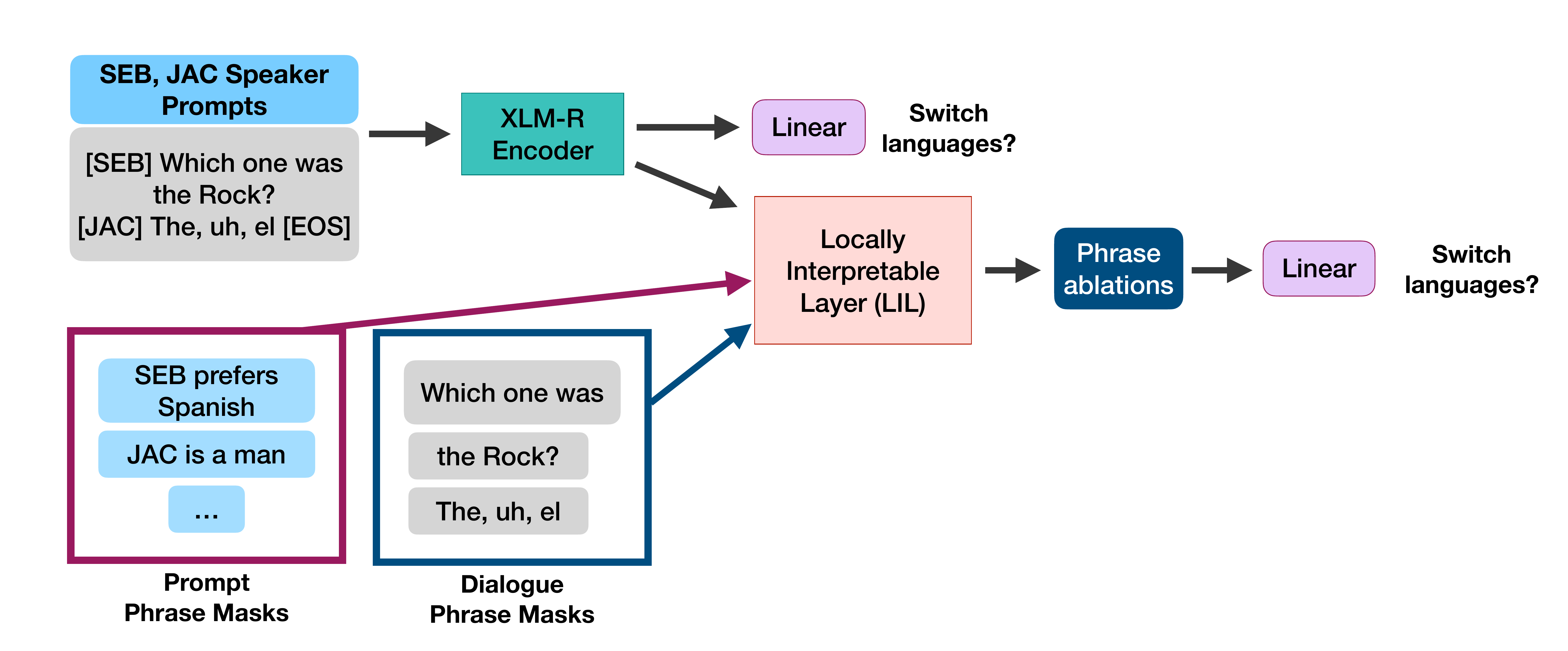}
    \caption{Architecture diagram of our proposed speaker-prompted code-switch prediction models. The input to the model is the dialogue context (gray) with descriptions of each speaker (blue) prepended to the dialogues. We encode the input using XLM-R (dark green) and use a linear layer (purple, top) to predict whether to code-switch or not. The encoded sentence, along with phrase masks, is passed to the Locally Interpretable Layer (LIL) \cite{rajagopal2021selfexplain}; using phrase ablation, LIL highlights influential phrases in the input by comparing local predictions to the full-sentence prediction. Baseline models follow a similar setup, but without any input from speaker prompts.}
    \label{fig:architecture}
    
\end{figure*}

Figure \ref{fig:architecture} illustrates the high-level model architecture of our proposed \textsc{sp-xlmr} models. The inputs include speaker information prepended to dialogue prompts and phrase masks for the LIL part of SelfExplain. Our baseline model, \textsc{xlmr} is similar, except it does not input speaker descriptions or speaker phrase masks.

\end{document}